\documentclass[sigplan,screen,10pt]{acmart}
\renewcommand\footnotetextcopyrightpermission[1]{}
\usepackage{xspace}
\newcommand{\AUTHORS}{Authors}
\newcommand{\NAME}{{\scshape Lynx}\xspace}
\newcommand{\TITLE}{\Large \NAME: Enabling Efficient MoE Inference Through Dynamic Batch-Aware Expert Selection}
\newcommand{\KEYWORDS}{}
\newcommand{\CONFERENCE}{}
\newcommand{\PAGENUMBERS}{yes}%
\newcommand{\COMMENTS}{no}%

\newcommand{\paraf}[1]{\noindent\textbf{#1}}
\usepackage{balance, ifthen, multirow}
\usepackage{times}
\usepackage[normalem]{ulem}

\usepackage[font=small, textfont=bf, labelfont=bf, aboveskip=0pt]{caption}
\usepackage{subcaption}

\usepackage{fancyhdr}

\ifthenelse{\equal{\PAGENUMBERS}{yes}}{%
  \pagestyle{plain}
}{%
  \pagestyle{empty}
}

\usepackage{enumitem}
\setlist{itemsep=0pt,parsep=0pt,topsep=0pt}%
\newcommand{\squishlist}{
   \begin{list}{$\bullet$}
    { \setlength{\itemsep}{0pt}      \setlength{\parsep}{3pt}
      \setlength{\topsep}{1pt}       \setlength{\partopsep}{0pt}
      \setlength{\leftmargin}{1.0em} \setlength{\labelwidth}{1em}
      \setlength{\labelsep}{0.5em} } }
\newcommand{\squishend}{
    \end{list}  }
\usepackage{booktabs}
\usepackage{colortbl}
\usepackage{float}                           %

\definecolor{placeholderbg}{rgb}{0.85,0.85,0.85}

\usepackage{mathtools}
\usepackage{pifont}

\usepackage{url}
\hypersetup{%
    pdfauthor = {\AUTHORS},
    pdftitle = {\TITLE},
    pdfsubject = {\CONFERENCE},
    pdfkeywords = {\KEYWORDS},
    bookmarksopen = {true}
}
\usepackage{cleveref}%
\crefname{section}{\S}{\SS}
\crefformat{section}{\S#2#1#3} %
\crefformat{subsection}{\S#2#1#3}
\crefformat{subsubsection}{\S#2#1#3}

\usepackage{listings}
\newcommand\code[1]{\lstinline$#1$}

\lstdefinelanguage{paper}{
 keywords={partition, transform, gather, scatter, apply},
 keywordstyle=\color{blue}\bfseries,
 morekeywords={[2]degrees,branch,commit,v_prev},
 keywordstyle={[2]\color{red}\bfseries},
 morekeywords={[3]if,def,Class,return,else,None,False,True,Array,while,G},
 keywordstyle={[3]\bfseries},
 basicstyle=\small\ttfamily,
 identifierstyle=\color{black},
 sensitive=false,
 comment=[l]{\/\/},
 morecomment=[s]{/*}{*/},
 commentstyle=\color{green}\ttfamily,
 stringstyle=\color{red}\ttfamily,
 breaklines=true,
}

\lstnewenvironment{python}
{\lstset{
    language=paper,
    frame=tlbr,
    columns=fullflexible,
    postbreak=\mbox{\textcolor{red}{$\hookrightarrow$}\space},
    captionpos=b}}
{}

\usepackage{algorithm}%
\usepackage{algpseudocode}%

\usepackage{tikz}%

\usepackage{savesym}
\savesymbol{comment}
\ifthenelse{\equal{\COMMENTS}{yes}}{%
    \setlength{\marginparwidth}{1.5cm}
    \usepackage[draft]{styles/commenting}
}{%
    \usepackage[final]{styles/commenting}
}%

\declareauthor{anand}{Anand}{magenta}

\declareauthor{ada}{Ada}{orange}

\declareauthor{vima}{Vima}{blue}

\declareauthor{jae}{Jae}{purple}

\declareauthor{john}{John}{olive}

\declareauthor{rohit}{Rohit}{teal}

\usepackage{soul}
\soulregister\cite7
\soulregister\ref7
\soulregister\pageref7

\usepackage[all]{nowidow}
\usepackage{amsmath}
\usepackage{xcolor}
\usepackage{graphicx}
\usepackage{subcaption}
\usepackage{microtype}
\usepackage{graphicx}
\usepackage{booktabs} %
\usepackage{xspace}
\usepackage{mdframed}
\usepackage{enumitem}
\usepackage{amsfonts}
\newmdenv[
    linewidth=1pt,
    topline=false,
    rightline=false,
    bottomline=false,
    leftmargin=10pt,
    rightmargin=10pt,
    backgroundcolor=gray!10,
    innerleftmargin=10pt,
    innerrightmargin=10pt,
    innertopmargin=5pt,
    innerbottommargin=5pt
]{takeaway}

\usepackage{hyperref}

\title{\TITLE}

\author{Vima Gupta}
\affiliation{\institution{Georgia Institute of Technology}\country{}}

\author{Jae Hyung Ju}
\affiliation{\institution{Georgia Institute of Technology}\country{}}

\author{Kartik Sinha}
\affiliation{\institution{Georgia Institute of Technology}\country{}}

\author{Ada Gavrilovska}
\affiliation{\institution{Georgia Institute of Technology}\country{}}

\author{Anand Iyer}
\affiliation{\institution{Georgia Institute of Technology}\country{}}

\begin{abstract}
Modern foundational models overwhelmingly adopt Mixture-of-Experts (MoE) architectures, prized
for their selective parameter activation that decouples parameter count from computational cost.
However, MoEs face a fundamental tension in serving: batching, critical for
throughput, forces the activation of many experts, negating MoEs' sparsity
benefits and making decode firmly memory-bandwidth-bound. We present \NAME, the first system---to the best of our
knowledge---that alleviates the memory bandwidth bottleneck in MoE inference in a workload-agnostic
fashion. \NAME leverages a key property of MoE training: load-balancing losses introduce batch-level expert
activation skews and redundancy, which it exploits by remapping low-affinity token-to-expert
assignments within each batch using a novel \emph{AffinityBinning} technique that reduces the total
experts invoked. We evaluate \NAME on state-of-the-art models from four families across eight benchmarks spanning
code generation, mathematics, reasoning, and vision tasks, where it achieves up to $1.30\times$
lower latency compared to the baselines. These improvements don't come at a loss: \NAME frequently \emph{improves} accuracy (on average)
and incurs only less than 1\% drop in the worst case, while being complementary to existing
techniques.
\end{abstract}

\begin{document}

\maketitle
\thispagestyle{plain}
\pagestyle{plain}

\ifthenelse{\equal{\PAGENUMBERS}{no}}{%
  \thispagestyle{empty}
}

\makeatletter
\def\blfootnote{\xdef\@thefnmark{}\@footnotetext}
\makeatother

\lstset{language=paper}
\sloppypar

\section{Introduction}\label{sec:introduction}

Mixture-of-Experts (MoE) has become the de facto architecture in modern foundational models,
powering state-of-the-art offerings from families such as Qwen~\cite{Yang2024Qwen2TR, qwen3},
Llama~\cite{llama4_2024}, Mixtral~\cite{Jiang2024MixtralOE}, and DeepSeek~\cite{Shao2024DeepSeekV2AS,
DeepSeekAI2024DeepSeekV3TR}. MoEs replace dense feed-forward layers with multiple specialized
sub-networks, or \emph{experts}, and a learned routing network that directs each input token to a
small subset of these experts~\cite{Shazeer2017OutrageouslyLN, Lepikhin2020GShardSG,
Hwang2022TutelAM}. This \emph{selective activation} decouples parameter count from computational
cost, enabling models to scale to hundreds of billions of parameters while activating only a small
fraction per input. For instance, Qwen2-57B-A14B-Instruct activates just 14 of its 57 billion
parameters per token~\cite{Yang2024Qwen2TR}. Thus, MoEs offer the compelling promise of inference cost 
proportional not to total model size, but to the far smaller activated footprint.

Realizing this promise in practice, however, requires navigating a fundamental tension between
latency and throughput inherent to production serving. Processing requests one at a time would
respect per-request latency but leaves GPU utilization dismal; batching requests together amortizes
fixed costs and improves throughput, but accumulates latency. This tension is not merely theoretical:
providers today charge $4$--$6\times$ more for low-latency ``priority'' APIs compared to
throughput-oriented serving~\cite{openai_api_pricing, anthropic_claude_pricing}, a pricing premium
that reflects the real cost of reserving capacity for small, latency-constrained batches. 
Consequently, production deployments must settle on a per-GPU batch size that is neither too small to amortize infrastructure costs, nor too large to violate latency SLOs.

This tension manifests disproportionately for MoEs compared to dense models. Because the router independently selects a different expert subset for each token, the set of experts that must be loaded from GPU high-bandwidth memory (HBM) at each layer grows with the composition of the batch, as well as its size.
At the moderate  batch sizes imposed by latency SLOs, this dynamic, data-dependent memory access pattern keeps MoE decode  firmly memory-bandwidth-bound: latency scales directly with the number of distinct experts 
activated  across the batch~(\S\ref{sec:Background}). Worse, even moderate batches suffice to activate nearly all experts: in Qwen2-57B-A14B-Instruct, where each token selects 8 of 64 experts, a batch of just 8 diverse requests is enough to saturate the entire expert pool~\cite{Yang2024Qwen2TR}. At that point, MoEs offer no sparsity benefit whatsoever: they activate as many or more parameters as a dense model of equivalent capacity\footnote{An MoE with E experts each with P parameters generally underperform a dense model with E $\times$ P parameters.\cite{Guo2025AdvancingES},\S\ref{sec:Background}}, while additionally bearing the cost of dynamic 
expert movement in the critical path of every decode iteration.

The natural approach to this is to reduce the volume of data movement---either by shrinking expert 
size or by reducing the number of experts fetched. Existing techniques pursue both directions through pruning, quantization, and expert clustering~\cite{Lu2024NotAE, dong2025domain, lee2024stun, 
huang2025mixture, Chen2025EACMoEEA, Li2023MergeTC, Yang2024MoEICM, Liu2024EfficientEP}, and are effective for workloads they are calibrated for. However, they share two fundamental limitations. First, they depend on extensive offline calibration that assumes expert redundancy at the workload level---an assumption that is increasingly fragile as modern MoE models grow more expressive and their serving workloads more diverse~(\S\ref{sec:Background}). Second, they permanently alter the model: experts are discarded, merged, or compressed at compile time and cannot be recovered at serving time, making them inflexible to workload changes. More recent dynamic approaches utilize calibration-based signals to reduce experts on a per-token level~\cite{Lu2024NotAE, Chen2025EACMoEEA, huang2025mixture, guo2024dynamic}, but since each token still independently selects from the full expert pool, batch-level expert utilization remains high
as batch size grows. Neither class of techniques resolves the fundamental tension: the bottleneck lives at the batch level, and neither operates there.

In this paper, we present \NAME, the first system---to the best of our knowledge---that alleviates
the memory bandwidth bottleneck in MoE inference in a workload-agnostic fashion. \NAME requires no
calibration data, makes no permanent changes to the model, and operates entirely from signals
already available at runtime. Its key insight stems from a crucial property of MoE training:
load-balancing losses encourage routers to distribute tokens broadly across experts, even when
confidence in secondary selections is low~\cite{eo-etal-2025-mixture, Guo2025AdvancingES}. While
this forced diversification is essential during training, \NAME observes that it creates
redundancy during inference at the batch level, where it manifests as skewed expert activations at
each layer in every forward pass~(\S\ref{sec:Background}).

\NAME exploits this redundancy through a set of principles guided by three observations we uncover
through exhaustive analysis across multiple model families (\S\ref{Sec:Design}). First, the
router's output confidence scores reliably identify which token-to-expert mappings are potentially
redundant and safe to reassign. Second, the sorted order of the router's expert selections provides
a direct signal for accuracy impact: top-ranked experts disproportionately determine output quality,
while lower-ranked selections among the top-$k$ exhibit high redundancy.
Third, expert sensitivity
differs drastically between the prefill and decode phases, with prefill demanding strict expert
fidelity and decode showing remarkable resilience to reassignment. Armed with these principles, \NAME 
\emph{remaps} low-affinity token-to-expert assignments within a batch onto experts the batch is already 
activating, reducing the total number of experts invoked at a batch level. Crucially, \NAME does this 
without discarding any experts permanently, and without changing the number of experts each token 
activates---preserving the top-$k$ activation semantics of the model\footnote{Since MoEs are 
trained to always use top-k, changing the k dynamically may not be accommodated by the model without 
post-training.~\cite{Yue2025AdaKRB}}.

Making remapping practical requires solving a non-trivial coupled optimization in the critical path
of inference: the question of whether a given expert can be dropped depends on whether \emph{every} 
token in the batch that relies on it can be safely redirected elsewhere, making per-token and batch-level
decisions inherently interdependent. \NAME solves this using its novel \emph{AffinityBinning} technique. 
AffinityBinning discretizes each token's router confidence scores into bins whose width and count 
are determined solely by the model's sparsity ratio\footnote{Ratio of how many ($k$) experts are activated 
to the total ($N$).}, not by the workload or the task. This makes \NAME self-calibrating: it adapts 
automatically to any MoE architecture without profiling or tuning, while the batch-size-adaptive scoring 
naturally adjusts expert retention pressure to the competition among tokens in each forward pass. \NAME 
implements AffinityBinning across three lightweight components: the \emph{confidence analyzer} identifies 
tokens whose expert assignments are amenable to remapping; the \emph{adaptive expert scorer} jointly 
determines, across all tokens in the batch, which experts can be eliminated and which must be retained; 
and the \emph{expert remapper} redirects low-affinity assignments to the surviving expert set.

We implement \NAME on vLLM~\cite{kwon2023efficient}, with the three components realized as fused
CUDA kernels that intercept the router output at every layer with negligible overhead even in
the critical path~(\S\ref{sec:implementation}). 
We evaluate \NAME on state-of-the-art models from four families, namely Qwen, DeepSeek, Mixtral, and Llama, across six benchmarks spanning code
generation, mathematics, reasoning and vision tasks. \NAME achieves up to $1.30\times$ lower latency  while being within 1\% accuracy loss across all benchmarks. Strikingly, \NAME frequently \emph{improves} accuracy over the baseline---a consequence of remapping low-confidence expert assignments that were forced by training-time load-balancing constraints rather than genuine token-expert affinity.
\NAME is complementary to,
not competing with, existing MoE optimizations: applied atop
state-of-the-art offloading and quantization techniques, it boosts their performance by up to 31\% and 10\%, respectively. Since \NAME relies solely on signals intrinsic to any MoE model, it generalizes across architectures without modification. As GPU compute continues to outpace memory bandwidth growth and the arithmetic intensity gap that makes MoE decode memory-bound widen, we believe that 
\NAME's techniques will become increasingly critical to realizing the efficiency promise of MoE architectures on future hardware.

\section{Motivation \& Challenges}\label{sec:Background}

  \begin{figure*}[!htb]
  \centering
  \includegraphics[width=0.7\textwidth]{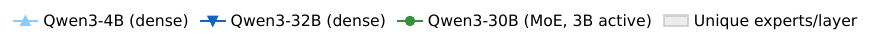}
    \\[2pt]
    \begin{subfigure}[b]{0.19\textwidth}
      \centering
      \includegraphics[height=1.4in]{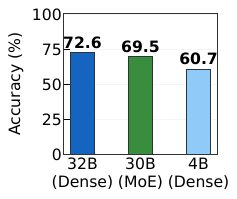}
      \caption{Model accuracy}
      \label{fig:qwen3-accuracy}
    \end{subfigure}
    \hspace{-2mm}
    \begin{subfigure}[b]{0.51\textwidth}
      \centering
      \includegraphics[height=1.4in]{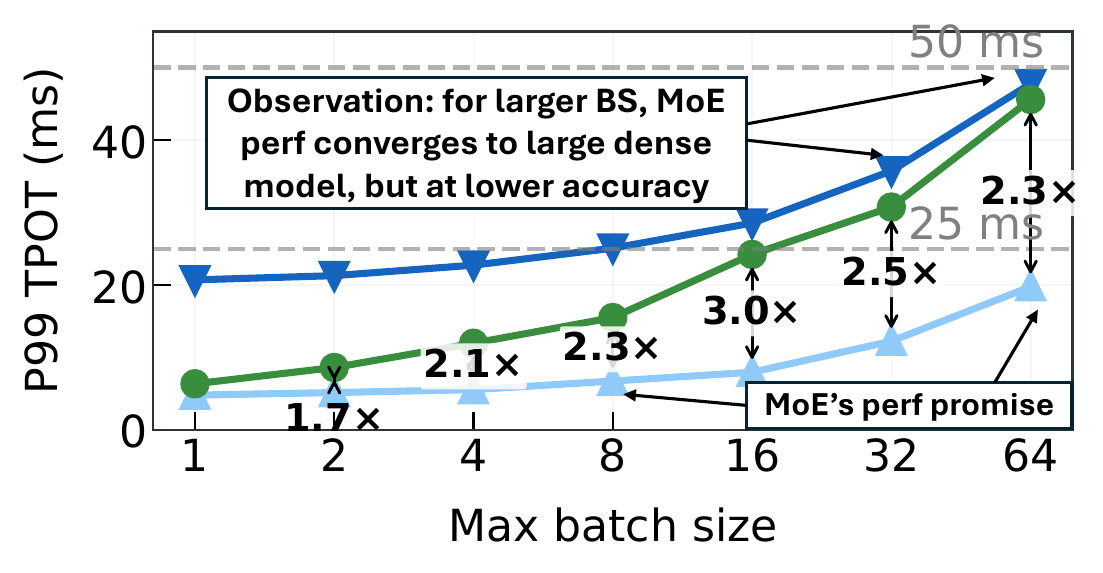}
      \caption{SLO-compliant region}
      \label{fig:qwen3-slo-region}
    \end{subfigure}
    \hspace{-7mm}
    \begin{subfigure}[b]{0.29\textwidth}
      \centering
      \includegraphics[height=1.4in]{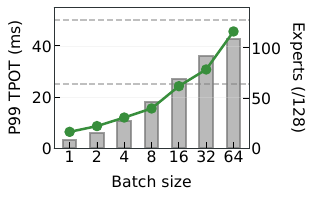}
      \caption{Expert diversity (Qwen3-30B)}
      \label{fig:qwen3-expert-diversity}
    \end{subfigure}
    \caption{Performance and accuracy of dense models (Qwen3-4B and Qwen3-32B) and MoE (Qwen3-30B-A3B): (a) MoE achieves accuracy of a 32B dense model with just 3B active parameters per input. (b) MoE performance degrades from being similar to 4B model to convering to 32B dense model as batch size increases. (c) Root cause is the increase in activated parameters at larger batches.}
    \label{fig:qwen3-bs-sweep}
  \end{figure*}

We begin with an overview of MoE architecture and how batching breaks the promise of MoEs, establishing why decode is
fundamentally memory-bandwidth-bound~(\S\ref{subsec1:Background}). We then highlight a key
property of MoE training that creates exploitable redundancy at the batch level during
inference~(\S\ref{subsec2:Background}). Finally, we describe the challenges in realizing these
benefits in practice, which motivate \NAME's design~(\S\ref{subsec:challenges}).
\subsection{Problem: Batching Breaks the MoE Promise}\label{subsec1:Background}
MoE architecture is a variant of dense LLM architecture, which replaces the dense feed-forward layers in each decoder block with $N$ specialized sub-networks
(\emph{experts}) and a learned router. For each input token, the router computes logits $z_i$,
applies softmax to produce a probability distribution $p_i = {e^{z_i}}/{\sum_{j=1}^N e^{z_j}}$
over all $N$ experts, and selects the top-$k$ experts per token for computation. 
This fraction of
experts activated per token, $k/N$, is a model architectural property known as the \emph{sparsity ratio}. Representative values are $0.25$ for
Mixtral-8x7B~\cite{Jiang2024MixtralOE} ($k{=}2, N{=}8$), $0.125$ for
Qwen-2~\cite{Yang2024Qwen2TR} ($k{=}8, N{=}64$), and $0.03$ for
DeepSeek-V3~\cite{DeepSeekAI2024DeepSeekV3TR} ($k{=}8, N{=}256$). While finding the optimal
sparsity ratio remains an active area of research~\cite{nakamura2026optimal, Abnar2025ParametersVF,
elango2026latentmoeoptimalaccuracyflop}, recent model families~\cite{llama4_2024,
Shao2024DeepSeekV2AS} have trended towards lower $k/N$ by increase $N$ while holding $k$ fixed. 

\paraf{MoE serving:} MoEs promise the accuracy of a large dense model while incurring only the inference cost of a small dense model with parameter count proportional to the MoE's per-input acitve parameters. We evaluate this claim by comparing Qwen3-30B-A3B (MoE) against Qwen3-4B (small dense model) and Qwen3-32B (large dense model) on a real-world trace, ShareGPT~\cite{sharegpt_vicuna_unfiltered}, on an H200 at different batch sizes.

At a batch size of 1, the MoE model keeps this promise: its decode latency matches the small dense model (Figure~\ref{fig:qwen3-slo-region}, left arrow), while achieving 8.8\% higher average accuracy across tasks~\cite{qwen3}. However, at higher batch sizes (Figure~\ref{fig:qwen3-slo-region}(b), right arrow), MoEs break this promise. Their latency approaches that of the large dense model, while the MoE average accuracy trails behind that of the large model's accuracy by 3.6\%.

\paraf{Batching and SLOs:} 
To study the impact of batching, we analyze the p99 latencies across different batch sizes in Figure~\ref{fig:qwen3-slo-region}.
The MoE model is consistently slower than the small dense model by 1.7$\times$ to 3$\times$ at p99 latency as batch size increases. Since each token independently selects its top-k experts, the number of activated experts grows with batch size, as confirmed by the linear correlation that we observe in Figure~\ref{fig:qwen3-expert-diversity} between batch size and the number of activated experts. Thus, while batching incurs only computational overhead in dense models, MoEs additionally incur data movement cost from activating more experts~\cite{elango2026latentmoeoptimalaccuracyflop, adhinarayanan2026qsinequalityquantifyingdouble}, paying a much steeper cost.

Production services rely on request batching to sustain throughput~\cite{Yu2022OrcaAD, kwon2023efficient}, but latency SLO constraints restrict the maximum batch size~\cite{Yun2024TowardIM}. In line with  prior work and deployed API services~\cite{openai_api_pricing}, we use two SLOs: 50\,ms (20 tokens/s/user) and 25\,ms (40 tokens/s/user), representing agents and chatbots~\cite{li2025adaserveacceleratingmultislollm, 10.5555/3691938.3691949}. We find that MoE data movement is difficult to overlap with the dynamic and fine-grained expert computation because expert activation patterns are irregular and input-dependent~\cite{zhang2025cometfinegrainedcomputationcommunicationoverlapping}. To meet the 25\,ms SLO, 42\% of the median decode iteration is spent fetching activated expert weights from HBM, creating a hard upper bound on achievable user throughput.

\paraf{Prefill Vs Decode}: MoEs process input in two phases, prefill and decode, where decode dominates serving costs at 2--8$\times$ the expense of input tokens~\cite{anthropic_claude_pricing, openai_api_pricing}. Each forward pass during auto-regressive decode consists of attention and expert layers. Attention operator latency grows linearly with batch size and quadratically with sequence length, motivating a rich body of optimization work on both kernels and model architectures~\cite{zadouri2026flashattention4algorithmkernelpipelining, Choudhary2025OptimizingAO, Zhu2024SampleAttentionNA, Sieber2024UnderstandingTD, DeepSeekAI2024DeepSeekV3TR}. While prior works on expert computation optimizations have similarly focused on compute-intensive training workloads~\cite{gale2022megablocksefficientsparsetraining,zhang2026moeblazebreakingmemorywall}, expert computation during inference has a different computational profile.

\begin{figure}[!t]
\centering
\includegraphics[width=0.8\columnwidth]{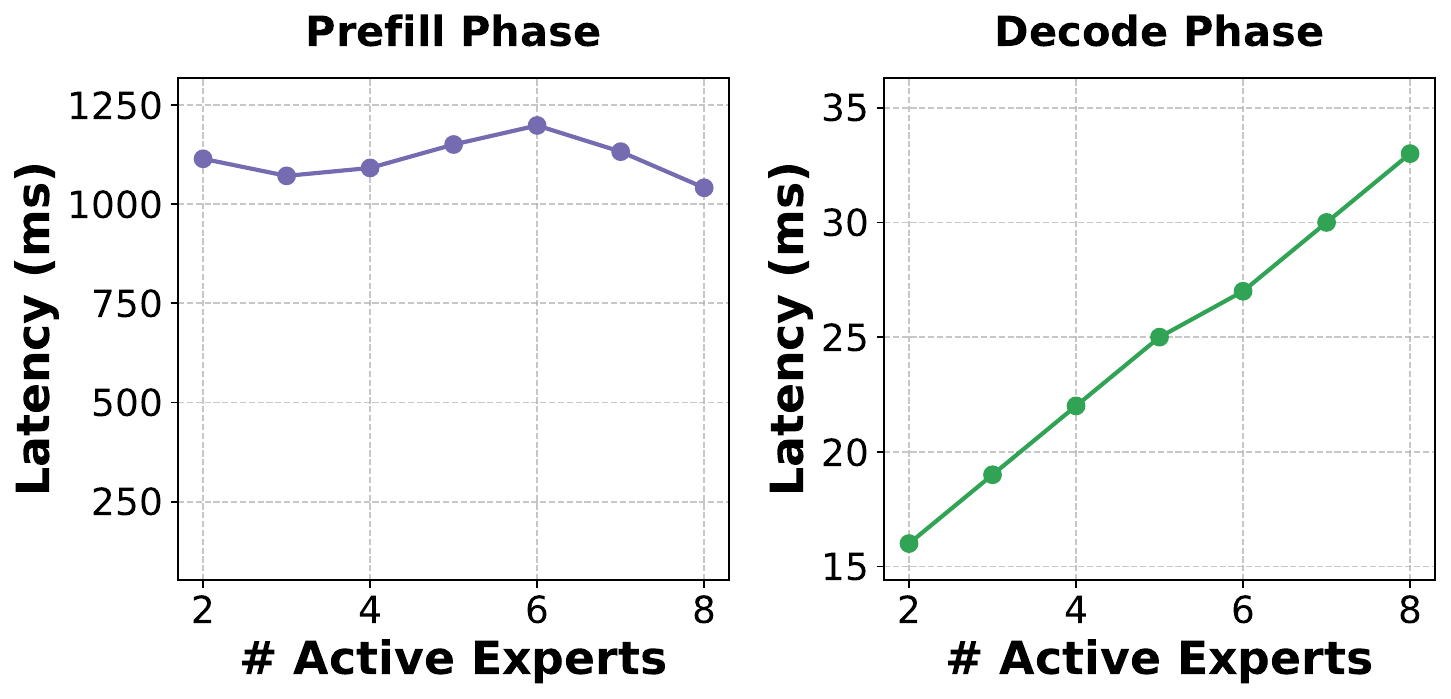}
\caption{Left: Prefill latency doesn't vary with active
experts (compute-bound). Right: Decode latency scales linearly with the number
of active experts (memory bandwidth bound)}
\label{fig:latency_trends}
\end{figure}

The arithmetic intensity of expert computation during inference is proportional to $\frac{n\cdot k}{N}$,
where $n$ is the number of tokens in the batch, $k$ is the number of tokens activated per input and  $N$ is the total number of experts. Figure~\ref{fig:latency_trends} showcases the arithmetic intensity discrepancy between the prefill and decode phase of inference. We run Mixtral-8x7B-v0.1~\cite{Jiang2024MixtralOE} on the ShareGPT dataset with 2 A100s and track the inference phase, activated experts and corresponding latency of the batch. We observe that prefill batches have high and constant latency even as the activated expert count varies, owing to compute-boundedness due to large number of tokens. Decode phase remains memory-bandwidth bound, owing to one token generated per request per iteration.

\begin{takeaway}
\textbf{Observation 1:} During the memory-bound decode phase, selective parameter activation is at
odds with batching under tight latency SLOs: decode latency scales with the total number of experts
activated across the batch, not per input.
\end{takeaway}

\subsection{Opportunity: Heterogeneity in Per-Batch Expert Activation Patterns}\label{subsec2:Background}
\begin{figure}[t!]
\centering
\includegraphics[width=0.8\linewidth]{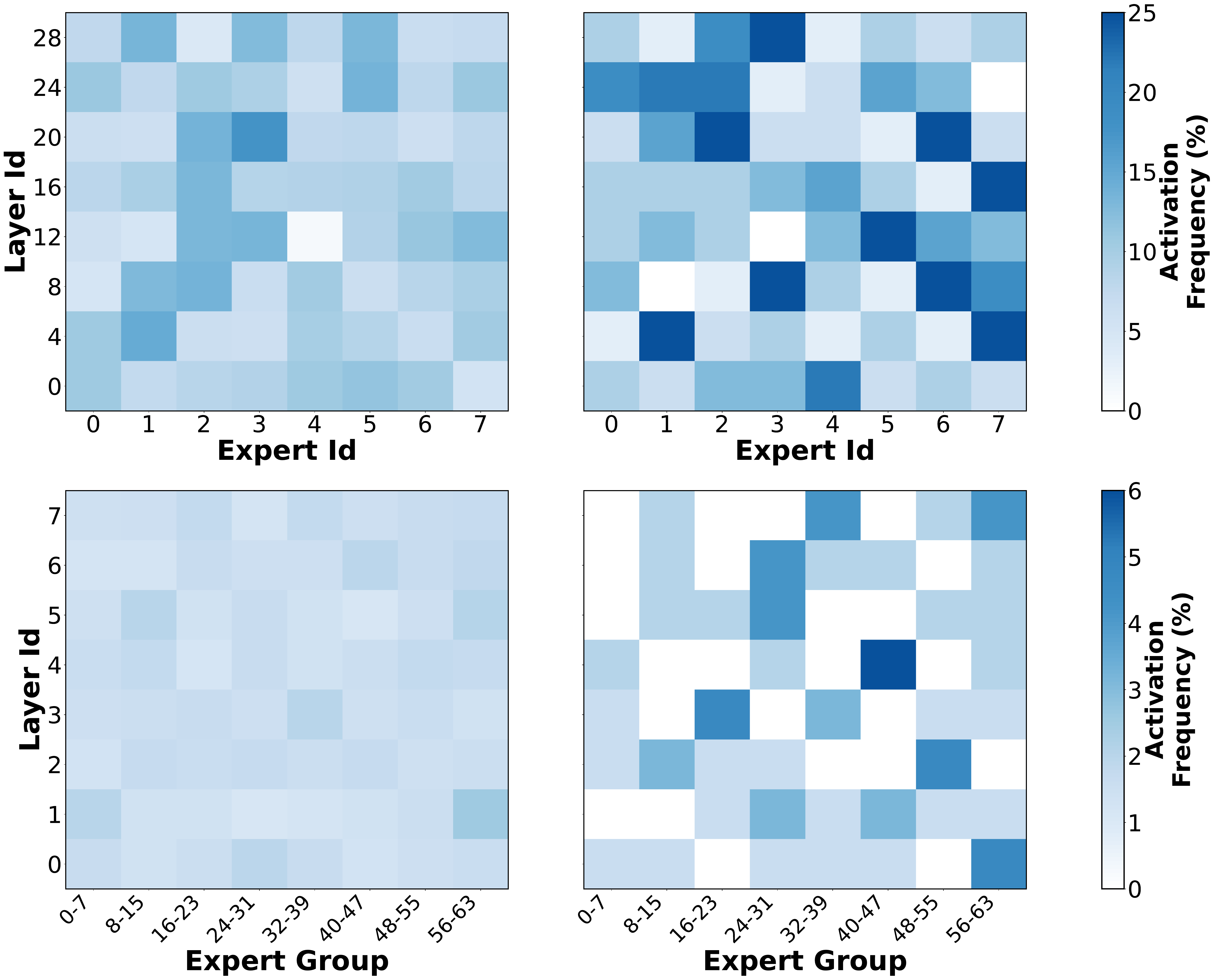}
\caption{Comparison of expert activation patterns at:
(Left) aggregate dataset-level - uniform;
(Right) batch-level - skewed (for Mixtral-8x7B (upper) and Qwen2 (lower)) for batch size=16.}
\label{fig:activation_patterns}
\end{figure}

Expert assignment for each token is determined at runtime by the router's softmax probabilities.
During training, an auxiliary load-balancing loss is applied to prevent expert collapse and promote
uniform expert utilization~\cite{Shazeer2017OutrageouslyLN, Lepikhin2020GShardSG,
Qiu2025DemonsIT}. This enforced uniformity persists at inference time.
Figure~\ref{fig:activation_patterns} (Left) analyzes expert activation frequency across millions
of tokens from the ShareGPT-V3~\cite{sharegpt_vicuna_unfiltered} dataset: at the workload level, no expert is consistently
underutilized~\cite{Lu2024NotAE, Jiang2024MixtralOE}. As a result, workload-level expert
redundancy is difficult to identify without a calibration dataset, rendering static expert
reduction techniques unsuitable for online serving~\cite{wu2026sere}.

However, the picture is substantially different at the batch level. In iteration-level scheduling,
which is standard in LLM inference systems~\cite{kwon2023efficient}, batch composition varies
between forward passes. This variation produces significant heterogeneity in which
experts are activated per iteration, even as the aggregate distribution remains uniform.
Figure~\ref{fig:activation_patterns} (Right) quantifies this for Mixtral-8x7B and Qwen2: mean
expert activation variability is $\sim$15--20\% at the batch level, compared to only 1.2\%
in aggregate. This batch-level skew suggests that dynamic, per-iteration expert management could
meaningfully reduce the number of experts fetched per decode step.

\begin{takeaway}
\textbf{Observation 2:} Expert activation exhibits strong heterogeneity at the batch level despite
remaining uniform in aggregate.
\end{takeaway}

This heterogeneity is a direct consequence of the load-balancing regularizer applied during
training. While essential for preventing expert collapse, the regularizer induces a systematic
side effect: the router learns to distribute tokens more broadly across experts than is strictly
necessary for output quality~\cite{eo-etal-2025-mixture, Qiu2025DemonsIT}. Specifically, when
the router's confidence is low---that is, when the gap between its highest- and lower-ranked
expert scores is small---secondary expert assignments are driven more by the training-time
diversity constraint than by genuine token-expert affinity. The result is a systematic gap between
experts activated out of \emph{necessity} and those activated as a byproduct of \emph{training
regularization}; it is precisely this gap that \NAME exploits.

\subsection{Challenges}\label{subsec:challenges}

The batch-level skew in expert activation creates a concrete opportunity: by identifying and
suppressing redundant expert activations within each batch, one can directly reduce the volume of
data transferred from HBM per decode iteration. Realizing this opportunity, however, requires
overcoming several non-trivial obstacles. Because the batch is recomposed at every iteration by
the continuous batching scheduler~\cite{kwon2023efficient}, any solution must operate entirely at
runtime, and adapt to the routing decisions at each layer, without prior knowledge of the workload. It must further generalize across heterogeneous
MoE architectures, including those with shared experts~\cite{Yang2024Qwen2TR}, expert
grouping~\cite{DeepSeekAI2024DeepSeekV3TR}, and widely varying sparsity ratios, while remaining
lightweight enough to execute in the critical path of inference. \NAME addresses three key
challenges in constructing such a system.

\smallskip
\noindent\textbf{Identifying Important Tokens.} Tokens within a batch differ substantially in
their sensitivity to expert reassignment. Assigning equal importance to every token's expert
selection would, even at modest batch sizes, result in retaining nearly the full expert set,
negating any bandwidth benefit. A more principled approach requires a notion of \emph{relative
importance}: certain tokens genuinely require their selected experts for accurate output, while
others can be redirected to alternative experts without loss of output
quality~\cite{jaiswal2025finding}. This distinction must be drawn without task-specific
thresholds or calibration data, and must remain valid across architectures with differing
sparsity ratios. This raises our first challenge: \emph{How can we identify tokens amenable to
expert remapping without incurring accuracy loss?}

\smallskip
\noindent\textbf{Selecting Critical Experts per Batch.} Identifying remappable tokens is
necessary but not sufficient; an expert may only be eliminated from the active set if it is
non-critical for \emph{every} token in the batch. An expert that is dispensable for one token
but essential for another cannot safely be dropped. The core difficulty is reconciling per-token
expert preferences, which vary widely in the strength of their conviction, into a single
batch-level decision on the active expert set. Naive voting schemes that tally top-$k$ selections
across the batch~\cite{Lu2024NotAE, Yue2025AdaKRB} treat all selections as equally weighted: a
weakly preferred expert of an indifferent token receives the same vote as the decisive top choice
of a high-conviction token. This preserves unnecessary experts in the active set while providing
no mechanism to redirect low-conviction tokens to equally viable alternatives. This raises our
second challenge: \emph{How can we determine the minimal critical expert set for a batch while
respecting the varying conviction of per-token routing decisions?}

\smallskip
\noindent\textbf{Opportunistic Expert Reduction.} Expert remapping yields latency benefits only
when the decode iteration is memory-bandwidth-bound. In compute-bound regimes, such as during
prefill or under high system load, reducing the expert set provides no throughput improvement and
introduces unnecessary routing overhead. Furthermore, sensitivity to expert reassignment varies
across layers within the same forward pass; certain layers tolerate aggressive reduction while
others do not. \NAME must therefore determine \emph{when} and \emph{where} remapping is
beneficial without recourse to offline profiling or workload-specific tuning. This raises our
third challenge: \emph{How can we reliably identify opportunities for expert remapping across
inference phases and model layers at runtime?}

\smallskip
Addressing these challenges requires a data-driven characterization of how expert
selection influences output accuracy across multiple granularities: individual tokens, batches,
layers, and full inference iterations. The following section presents the empirical observations
that ground \NAME's design and describes how each challenge is resolved.

\begin{figure}
\centering
\includegraphics[width=0.7\linewidth]{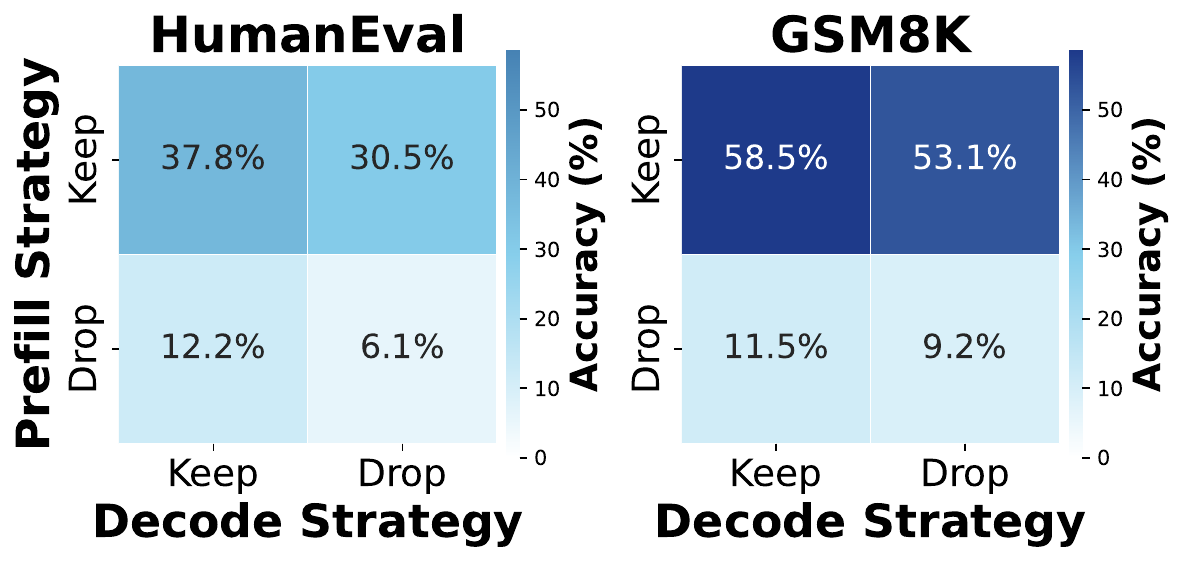}
\caption{Phase-specific impact of expert reassignment. Code generation (HumanEval) and reasoning (GSM8K) are sensitive to expert reassignment during prefill but resilient during decode, revealing an asymmetry in MoE processing across phases.}
\label{fig:phase_sensitivity}
\end{figure}

\section{\NAME Design}
\label{Sec:Design}

\begin{figure*}[t]
  \centering
  \includegraphics[width=0.95\textwidth]{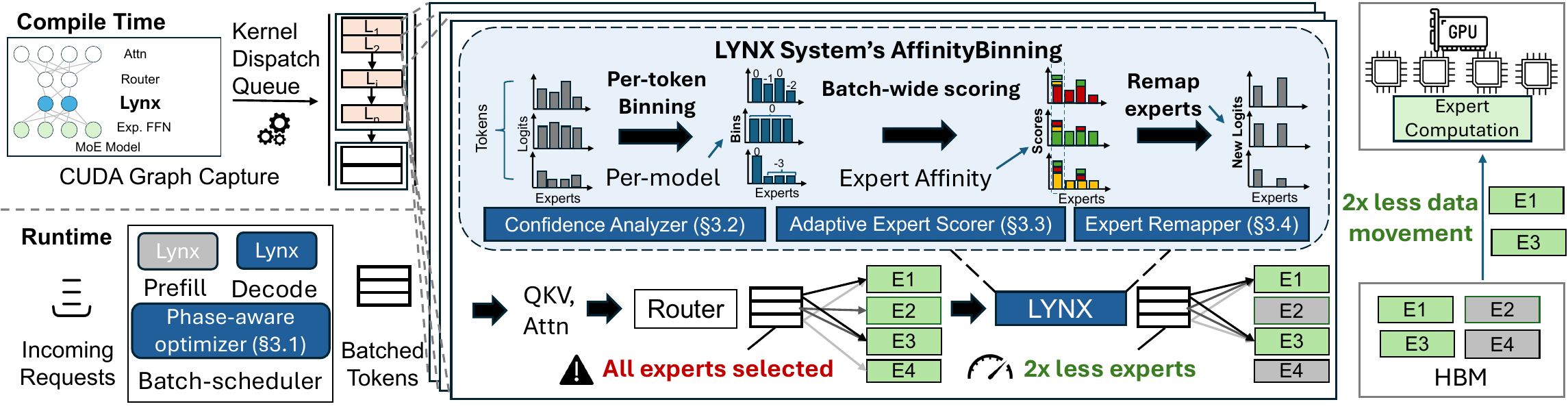}
  \caption{System architecture of \NAME. Left: the phase-aware optimizer identifies
  memory-bound inference phases; layer-level components are compiled using CUDA Graphs for
  maximum performance. Center: \NAME bins expert choices
  per token, computes batch-wide scores for each expert, selects the critical expert set, and
  remaps all token-to-expert assignments accordingly. Right: \NAME activates significantly fewer experts per batch, reducing
  memory traffic (two out of four activated in this example).}
  \label{fig:arch}
\end{figure*}

\NAME is a system that integrates with any LLM serving engine to mitigate the memory bandwidth
bottleneck in MoEs and improve serving performance. Its design is grounded in three empirical observations,
each of which resolves one of the challenges identified in~\S\ref{subsec:challenges} and
corresponds to a dedicated system component. First, the phase of inference --- prefill versus
decode --- determines whether expert remapping yields any benefit at all, motivating a
\emph{phase-aware optimizer} that applies \NAME's techniques selectively
(\S\ref{subsec:Phase}). Second, router confidence scores reliably distinguish tokens whose
expert assignments are critical from those amenable to remapping, motivating a \emph{confidence
analyzer} (\S\ref{subsec:Conf}). Third, the ranked order of expert selections reveals that
top-ranked experts dominate output quality while lower-ranked selections are highly redundant,
motivating an \emph{adaptive expert scorer} (\S\ref{subsec:Adap}). Together, these components
implement \NAME's core operation: the \emph{expert remapper} consolidates token-to-expert
assignments batch-wide onto a minimal expert set (\S\ref{subsec:Remap}), and the
\emph{architecture and workflow} section describes how all components interact at runtime
(\S\ref{subsec:workflow}). We illustrate these observations using Qwen2-57B-A14B,
DeepSeek-V2-Lite-Chat, and Mixtral-8x7B-v0.1 with GSM8K and HumanEval as benchmarks.

\subsection{Phase Sensitivity}\label{subsec:prefill}

The first question \NAME must answer before applying any remapping is whether doing so will
yield a latency benefit. As established in~\S\ref{subsec1:Background}, expert remapping reduces
memory bandwidth consumption only when the decode iteration is memory-bandwidth-bound. Intervening
during compute-bound phases, such as prefill, would introduce routing overhead without any
corresponding benefit. Beyond this coarse phase distinction, sensitivity to expert reassignment
also varies across the two phases in a deeper sense that motivates different treatment.

Figure~\ref{fig:phase_sensitivity} quantifies this asymmetry. Expert reassignment
during prefill substantially degrades model performance on both code generation (HumanEval) and
complex reasoning (GSM8K), while similar modifications during decode produce minimal accuracy
impact across all task types. The consistency of this pattern across workloads of different
character suggests a fundamental property of autoregressive inference: the prefill phase
establishes the context that guides all subsequent computation, making it sensitive to expert
fidelity, while the decode phase benefits from complementary mechanisms---attention, residual
connections, and accumulated context---that compensate for suboptimal expert selection.

\begin{takeaway}
\textbf{Insight 1:} MoEs exhibit fundamentally different sensitivity to expert selection during
prefill versus decode phases, enabling phase-specific optimization strategies.
\label{Insight:phase_hierarchy}
\end{takeaway}

\subsubsection{Phase-aware optimizer}\label{subsec:Phase}
The phase-aware optimizer works in concert with the batch scheduler to determine whether a given
batch warrants \NAME's intervention. Its behavior adapts to three common serving policies:
\squishlist
    \item \emph{Co-located} deployments, where prefill and decode batches are scheduled on the
    same instance but not mixed: the optimizer identifies decode-only batches as memory-bound and
    sets a flag consumed by downstream components.
    \item \emph{Disaggregated} serving~\cite{10.5555/3691938.3691949}, where prefill and decode
    operations run on separate instances within the same replica: the optimizer marks the decode
    instance as memory-bound, requiring no per-batch classification.
    \item \emph{Chunked prefill} policies~\cite{10.5555/3691938.3691945}, where a single batch may
    contain both prefill and decode tokens: the optimizer labels batches consisting exclusively of
    decode tokens as memory-bound. Batches with a prefill chunk small enough to be memory-bound
    are a known edge case; addressing them risks degrading time-to-first-token (TTFT) and is left
    to future work.
\squishend
In all cases, the optimizer ensures that \NAME's remapping techniques are applied only where they
yield measurable benefit, with no intervention during compute-bound phases.

\subsection{Token Importance}\label{subsec:confidence}
Even within memory-bound decode iterations, not all tokens are equally amenable to expert
remapping. Reassigning the expert of a token with strong router conviction risks a meaningful
accuracy drop, while reassigning a token with weak conviction typically does not. \NAME exploits
this by distinguishing high-conviction tokens --- whose expert assignments are preserved --- from
low-conviction tokens --- whose assignments may be remapped.

\begin{figure}[t!]
      \centering
      \includegraphics[width=0.8\columnwidth]{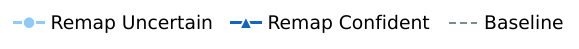}
      \vspace{2mm}
      \begin{subfigure}{0.48\columnwidth}
          \centering
          \includegraphics[width=\textwidth]{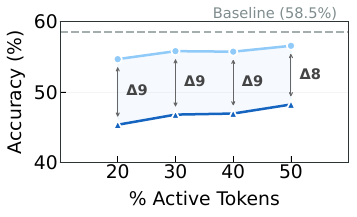}
          \caption{Mixtral}
          \label{fig:confidence-mixtral}
      \end{subfigure}
      \hfill
      \begin{subfigure}{0.48\columnwidth}
          \centering
          \includegraphics[width=\textwidth]{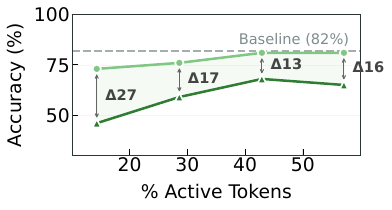}
          \caption{Qwen}
          \label{fig:confidence-qwen}
      \end{subfigure}
      \caption{Impact of token confidence on accuracy under selective expert remapping.
      Reassigning low-confidence tokens incurs substantially less accuracy degradation than
      reassigning high-confidence tokens.}
      \label{fig:token_importance}
\end{figure}

Figure~\ref{fig:token_importance} quantifies this relationship. We measure router confidence per
token as the variance in normalized probability scores, and vary the threshold above which tokens
are classified as high-confidence. As the threshold increases, a growing divergence emerges
between the accuracy impact of reassigning high- versus low-confidence tokens: high-confidence
tokens are consistently more sensitive to reassignment, while low-confidence tokens tolerate it
with minimal accuracy degradation. This divergence holds across model families, suggesting that
router logits contain a reliable signal for identifying critical token-expert mappings.

\begin{takeaway}
\textbf{Insight 2:} Router confidence provides a reliable signal for identifying critical
token-expert mappings; reassigning low-confidence mappings has minimal impact on downstream
accuracy.
\label{Insight:token_hierarchy}
\end{takeaway}

\subsubsection{Confidence analyzer}\label{subsec:Conf}
The confidence analyzer implements \NAME's \emph{AffinityBinning} technique, which
discretizes each token's router confidence into bins whose structure is determined solely by the
model's sparsity ratio --- not by any workload or task. This per-architecture calibration is
what makes \NAME self-calibrating and workload-agnostic.

Concretely, the confidence analyzer intercepts the router's probability weights at every MoE
layer. For each token, it measures the affinity of each expert relative to the top-1 expert by
computing the log-ratio of their softmax probabilities --- equivalently, the difference of the
corresponding logits before softmax.
These log-ratio values are then discretized into bins: a
value of 0 indicates highest affinity to the top-1 expert, and the bin index becomes increasingly
negative as the affinity gap widens (Figure~\ref{fig:arch}, right).
For routers that do not use softmax, e.g. sigmoid-based routers, we simply use the difference between the pre-sigmoid scores instead of log-ratio.

Two architecture-specific parameters, $\alpha$ and  $\beta$, control
the binning resolution: the width of each bin is inversely proportional to $\alpha$, and $\beta$
bounds the number of bins. Both are set once per model architecture based on its sparsity ratio
$(k/N)$, requiring no task-specific tuning. Among several candidate confidence measures such as variance, entropy, the Gini index, the log-ratio to the top-1 expert proved
most intuitive and empirically most representative.

\subsection{Expert Rank Hierarchy}\label{subsec:adaptive}

Knowing which tokens can be remapped is necessary but insufficient: the remapping must be
coordinated across the entire batch to ensure that no expert essential to any token is eliminated.
This requires understanding which experts, among those selected by a given batch, are truly
critical and which are redundant.

\begin{figure}[!t]
\centering
    \includegraphics[width=0.7\columnwidth]{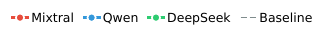}
    \centering
    \begin{subfigure}[b]{0.48\columnwidth}
      \centering
      \includegraphics[width=\textwidth]{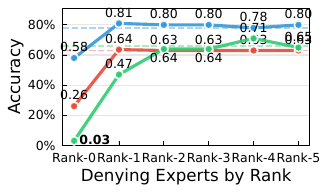}
      \caption{GSM8K benchmark}
      \label{fig:expert_denial_gsm8k}
    \end{subfigure}
    \hfill
    \begin{subfigure}[b]{0.48\columnwidth}
      \centering
      \includegraphics[width=\textwidth]{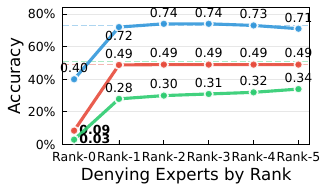}
      \caption{HumanEval benchmark}
      \label{fig:expert_denial_humaneval}
    \end{subfigure}
    \vspace{0.2cm}
    \caption{Impact of denying experts by rank on model accuracy. The top-ranked expert (E0) is
    critical across all model families; lower-ranked experts are substantially more redundant.}
    \label{fig:expert_denial}
\end{figure}

\begin{figure}[!t]
    \centering
    \includegraphics[width=0.6\columnwidth]{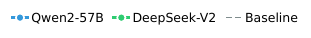}
    \begin{subfigure}[b]{0.48\columnwidth}
      \centering
      \includegraphics[width=\textwidth]{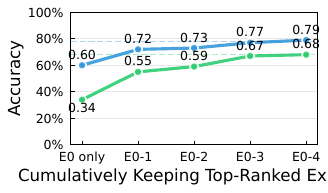}
      \caption{GSM8K benchmark}
      \label{fig:expert_keeping_gsm8k}
    \end{subfigure}
    \hfill
    \begin{subfigure}[b]{0.48\columnwidth}
      \centering
      \includegraphics[width=\textwidth]{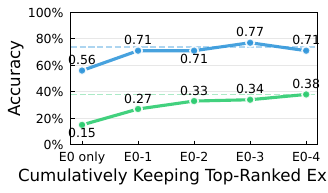}
      \caption{HumanEval benchmark}
      \label{fig:expert_keeping_humaneval}
    \end{subfigure}
    \vspace{0.2cm}
    \caption{Accuracy when cumulatively retaining top-ranked experts. Most performance is
    recovered after keeping 3--4 experts, with diminishing returns thereafter.}
    \label{fig:expert_keeping}
\end{figure}

\begin{figure}[t!]
      \centering
      \includegraphics[width=0.9\columnwidth]{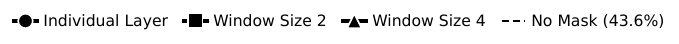}
      \includegraphics[width=0.75\columnwidth]{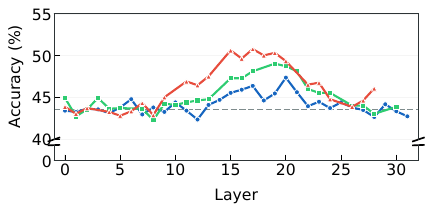}
      \caption{Layer-wise sensitivity analysis. Accuracy varies significantly across layers when
      expert assignments are selectively preserved, motivating layer-adaptive remapping policies.}
      \label{fig:layer}
\end{figure}

Figures~\ref{fig:expert_denial} and~\ref{fig:expert_keeping} reveal a striking asymmetry in
expert importance. Denying the top-ranked (rank-0) expert catastrophically degrades accuracy
across all model families and benchmarks, while denying lower-ranked experts causes minimal
disruption. Conversely, retaining only the top-ranked expert recovers most of the baseline
accuracy, with diminishing returns as additional experts are cumulatively restored. This hierarchy
is consistent across GSM8K and HumanEval, tasks of substantially different character, suggesting it reflects a structural property of MoE computation rather than a task-specific
artifact. Layer-wise analysis (Figure~\ref{fig:layer}) also shows variable sensitivity to
expert reassignment across layers: certain layers tolerate aggressive reduction while
others require expert fidelity, enabling layer-adaptive remapping policies.

\begin{takeaway}
\textbf{Insight 3:} The top-ranked expert dominates output quality; lower-ranked experts among
the top-$k$ exhibit high redundancy, and sensitivity varies across layers.
\label{Insight:expert_hierarchy}
\end{takeaway}

\subsubsection{Adaptive expert scorer}\label{subsec:Adap}
This component receives as input the AffinityBinning output from the confidence
analyzer, the per-token router probability scores, and the memory-bound flag from the phase-aware
optimizer. Its objective is to determine the minimal set of experts to activate for the batch
such that accuracy loss is minimized across all tokens.

To achieve this, the scorer implements a batch-wide scoring algorithm. For each expert, a
weighted score is added to its router logit based on how frequently other tokens in the batch
preferred it. Critically, rather than using vanilla weighted averaging, \NAME uses the batch size
as the base of an exponential weighting scheme and raises it to the bin value assigned to each
expert by the confidence analyzer. This formulation naturally adapts to competition among tokens
in the batch: experts preferred by many high-conviction tokens receive disproportionately higher
scores, while experts favored only by low-conviction tokens are down-weighted. The scorer
dynamically determines the final number of activated experts based on the resulting score
distribution, the bin width, and the number of allowed bins. High-confidence tokens are always
mapped to their top-ranked expert to ensure minimal accuracy impact.

For MoE architectures with shared experts such as DeepSeek-V2 and V3~\cite{DeepSeekAI2024DeepSeekV3TR,
Shao2024DeepSeekV2AS}, those experts are activated unconditionally at every layer, \NAME ensures these
are always included in the final expert set with a probability score of 1, preserving their
intended semantics. 

\subsection{Expert Remapper}\label{subsec:Remap}

The expert remapper is the final stage of \NAME's per-layer pipeline. It receives the minimal
expert set identified by the adaptive expert scorer and redirects low-confidence token-to-expert
assignments to experts within that set. The reassigned probability scores 
determine the final token-to-expert mapping, after which the component dispatches the expert
computation kernel with the appropriate kernel launch parameters.

Crucially, \NAME's remapping preserves the top-$k$ activation semantics of the underlying model:
every token still activates exactly $k$ experts, and no expert is permanently discarded. The
reduction in memory bandwidth comes entirely from consolidating the batch's expert activations
onto a smaller set, not from reducing per-token computation. This distinguishes \NAME from prior
approaches that reduce $k$ per token~\cite{Lu2024NotAE, Yue2025AdaKRB}, which alter model
semantics and require post-training to remain accurate. \NAME requires neither model modification
nor offline tuning, and adapts to any MoE architecture out of the box.

\subsection{Architecture and Workflow}\label{subsec:workflow}

Figure~\ref{fig:arch} presents the system architecture of \NAME and illustrates how its
components interact at runtime. The phase-aware optimizer is integrated within the batch
scheduler of the serving engine. The remaining components---confidence analyzer, adaptive
expert scorer, and expert remapper---operate at per-layer granularity within the model's
forward pass.

For each incoming batch, the phase-aware optimizer first determines whether the iteration is
memory-bound. If the batch is compute-bound, as during prefill, \NAME's components are bypassed
entirely for that iteration. For memory-bound decode batches, the MoE router at each layer
produces expert selection logits, which are intercepted by the confidence analyzer. The
confidence analyzer applies AffinityBinning to produce per-token bin assignments. The adaptive
expert scorer then computes batch-wide expert scores and determines the final active expert set.
Finally, the expert remapper redirects low-confidence token assignments to the active set and
dispatches the expert computation kernel. This workflow operates entirely within the critical
path of inference, requiring no model modifications or offline calibration.

\section{Implementation}\label{sec:implementation}
Our implementation works as a plug-and-play on top of vLLM~\cite{kwon2023efficient} through highly optimized CUDA graph compatible kernels in the critical path of MoE inference.  
\NAME's core  components: confidence analyzer, adaptive expert scorer and expert remapper  are implemented using four fused Triton~\cite{tillet2019triton} kernels while the phase-aware optimizer~\cref{subsec:Phase} operates within the batch scheduler.
Since we do not depend on any framework specific operations, \NAME can integrate with any LLM serving engine.

Each kernel in \NAME fuses operations that would otherwise launch over 700 PyTorch kernels, eliminating intermediate tensor data movement and excess memory traffic by keeping computations in registers or shared memory. This design reduces memory accesses, cuts kernel launch overhead, and maintains static control flow for efficient CUDAGraph capture. The routing policy is implemented in four steps: token-wise binning, batch-wise scoring, expert pruning, and expert remapping. The first kernel discretizes logits and computes top-k weight sums (per-token), the next two score and prune experts (per-batch) before recomputing softmax and top-\emph{k}, and the final kernel compacts the active-expert list, assigns tokens, and renormalizes weights. These four kernels replace over 700 of small PyTorch operations with four efficient launches~(Figure~\ref{fig:comp-breakdown}).

Current inference engines deploy  optimizations such as chunked prefill~\cite{10.5555/3691938.3691945}, disaggregated serving~\cite{10.5555/3691938.3691949}, CUDA Graphs, fused expert kernels with architecture-specific dispatch parameters during critical path of MoE inference. \NAME works seamlessly on top of all such SOTA optimizations to provide  performance benefits for MoE inference. \NAME is enabled through a CLI flag and automatically adapts to different number of active experts, batch size and available hardware, delivering immediate performance gains across diverse serving environments.

\begin{table}[!t]
\small
\centering
\caption{Models used to evaluate \NAME across families.}
\begin{tabular}{@{}ll@{}}
\toprule
\textbf{Model Family} & \textbf{Model} \\
\midrule
\multirow{3}{*}{Qwen}
& Qwen2-57B-A14B-Instruct~\cite{Yang2024Qwen2TR} \\
& Qwen3-30B-A3B-Instruct~\cite{qwen3} \\
& Qwen3-235B-A22B-Thinking-2507~\cite{qwen3} \\
\midrule
Mixtral
& Mixtral-8x7B-Instruct-v0.1~\cite{Jiang2024MixtralOE} \\
\midrule
DeepSeek
& DeepSeek-V2-Coder~\cite{Shao2024DeepSeekV2AS1} \\
\midrule
\multirow{2}{*}{Llama}
& Llama-4-Maverick-17B-128E-Instruct~\cite{llama4_2024} \\
& Llama-4-Scout-17B-16E-Instruct~\cite{llama4_2024} \\
\bottomrule
\end{tabular}
\label{tab:models-used}
\end{table}

\section{Evaluation}\label{sec:evaluation}

\begin{table*}[t]
  \centering
  \small
  \setlength{\tabcolsep}{3.5pt}
  \renewcommand{\arraystretch}{1.15}

\caption{\textsc{Lynx} achieves same accuracy (\colorbox{gray!85}{\textsc{\color{white}$\pm$X\%}}) and lower latency (\colorbox{green!8}{$\times$}) across code and math benchmarks on MoE models. The $\Delta$ column shows accuracy change on benchmark rows and TPOT reduction on green rows. Light gray rows report median and mean TPOT in milliseconds (Base $\to$ \textsc{Lynx}). The mean TPOT includes chunked prefill iterations. HE/MBPP use Pass@1, while MATH/GSM8K use EM (Exact Match) for accuracy measurement.}

  \begin{tabular*}{\textwidth}{@{\extracolsep{\fill}} l ccc ccc ccc ccc @{}}
  \toprule
  & \multicolumn{3}{c}{\textbf{Qwen2-57B-A14B-Instruct}}
  & \multicolumn{3}{c}{\textbf{Qwen3-30B-A3B-Instruct}}
  & \multicolumn{3}{c}{\textbf{DeepSeek-Coder-V2}}
  & \multicolumn{3}{c}{\textbf{Mixtral-8x7B-Instruct}} \\
  \cmidrule(lr){2-4} \cmidrule(lr){5-7} \cmidrule(lr){8-10} \cmidrule(lr){11-13}
  & Base & \textsc{Lynx} & $\Delta$
  & Base & \textsc{Lynx} & $\Delta$
  & Base & \textsc{Lynx} & $\Delta$
  & Base & \textsc{Lynx} & $\Delta$ \\
  \midrule

  \multicolumn{13}{l}{\textit{Code}} \\

  \rowcolor{gray!85}
  \quad \textsc{\color{white}he}
    & {\scriptsize \color{white}65.9\%} & {\scriptsize \color{white}65.9\%} & \color{white}$+$0.0\%
    & {\scriptsize \color{white}76.2\%} & {\scriptsize \color{white}75.2\%} & \color{white}$-$1.0\%
    & {\scriptsize \color{white}76.8\%} & {\scriptsize \color{white}78.7\%} & \color{white}$+$1.9\%
    & {\scriptsize \color{white}49.4\%} & {\scriptsize \color{white}50.6\%} & \color{white}$+$1.2\% \\
  \rowcolor{gray!6}
  \quad p50\,(ms)
    & {\scriptsize 15.83} & {\scriptsize 12.63} & \cellcolor{green!8}1.25$\times$
    & {\scriptsize 11.57} & {\scriptsize 9.51} & \cellcolor{green!8}1.22$\times$
    & {\scriptsize 28.34} & {\scriptsize 23.21} & \cellcolor{green!8}1.22$\times$
    & {\scriptsize 13.48} & {\scriptsize 11.45} & \cellcolor{green!8}1.18$\times$ \\
  \rowcolor{gray!6}
  \quad mean\,(ms)
    & {\scriptsize 27.83} & {\scriptsize 25.85} & \cellcolor{green!8}1.08$\times$
    & {\scriptsize 21.82} & {\scriptsize 19.26} & \cellcolor{green!8}1.13$\times$
    & {\scriptsize 34.26} & {\scriptsize 31.11} & \cellcolor{green!8}1.10$\times$
    & {\scriptsize 13.84} & {\scriptsize 12.19} & \cellcolor{green!8}1.14$\times$ \\

  \rowcolor{gray!85}
  \quad \textsc{\color{white}mbpp}
    & {\scriptsize \color{white}59.6\%} & {\scriptsize \color{white}60.4\%} & \color{white}$+$0.8\%
    & {\scriptsize \color{white}73.6\%} & {\scriptsize \color{white}74.0\%} & \color{white}$+$0.4\%
    & {\scriptsize \color{white}78.8\%} & {\scriptsize \color{white}78.8\%} & \color{white}$+$0.0\%
    & {\scriptsize \color{white}37.2\%} & {\scriptsize \color{white}37.6\%} & \color{white}$+$0.4\% \\
  \rowcolor{gray!6}
  \quad p50\,(ms)
    & {\scriptsize 16.03} & {\scriptsize 12.64} & \cellcolor{green!8}1.27$\times$
    & {\scriptsize 11.24} & {\scriptsize 9.45} & \cellcolor{green!8}1.19$\times$
    & {\scriptsize 28.35} & {\scriptsize 23.63} & \cellcolor{green!8}1.20$\times$
    & {\scriptsize 14.59} & {\scriptsize 13.34} & \cellcolor{green!8}1.09$\times$ \\
  \rowcolor{gray!6}
  \quad mean\,(ms)
    & {\scriptsize 18.79} & {\scriptsize 16.05} & \cellcolor{green!8}1.17$\times$
    & {\scriptsize 18.35} & {\scriptsize 16.91} & \cellcolor{green!8}1.09$\times$
    & {\scriptsize 39.08} & {\scriptsize 36.25} & \cellcolor{green!8}1.08$\times$
    & {\scriptsize 16.75} & {\scriptsize 15.43} & \cellcolor{green!8}1.09$\times$ \\

  \addlinespace[2pt]
  \multicolumn{13}{l}{\textit{Math}} \\

  \rowcolor{gray!85}
  \quad \textsc{\color{white}math}
    & {\scriptsize \color{white}47.1\%} & {\scriptsize \color{white}46.8\%} & \color{white}$-$0.3\%
    & {\scriptsize \color{white}64.4\%} & {\scriptsize \color{white}63.6\%} & \color{white}$-$0.8\%
    & {\scriptsize \color{white}11.2\%} & {\scriptsize \color{white}11.2\%} & \color{white}$+$0.0\%
    & {\scriptsize \color{white}37.8\%} & {\scriptsize \color{white}38.6\%} & \color{white}$+$0.8\% \\
  \rowcolor{gray!6}
  \quad p50\,(ms)
    & {\scriptsize 16.02} & {\scriptsize 12.57} & \cellcolor{green!8}1.27$\times$
    & {\scriptsize 11.55} & {\scriptsize 9.73} & \cellcolor{green!8}1.19$\times$
    & {\scriptsize 28.52} & {\scriptsize 23.62} & \cellcolor{green!8}1.21$\times$
    & {\scriptsize 14.54} & {\scriptsize 13.38} & \cellcolor{green!8}1.09$\times$ \\
  \rowcolor{gray!6}
  \quad mean\,(ms)
    & {\scriptsize 16.83} & {\scriptsize 13.70} & \cellcolor{green!8}1.23$\times$
    & {\scriptsize 14.11} & {\scriptsize 12.26} & \cellcolor{green!8}1.15$\times$
    & {\scriptsize 31.44} & {\scriptsize 27.59} & \cellcolor{green!8}1.14$\times$
    & {\scriptsize 19.42} & {\scriptsize 14.67} & \cellcolor{green!8}1.32$\times$ \\

  \rowcolor{gray!85}
  \quad \textsc{\color{white}gsm8k}
    & {\scriptsize \color{white}78.0\%} & {\scriptsize \color{white}77.2\%} & \color{white}$-$0.8\%
    & {\scriptsize \color{white}88.4\%} & {\scriptsize \color{white}87.6\%} & \color{white}$-$0.8\%
    & {\scriptsize \color{white}63.2\%} & {\scriptsize \color{white}64.0\%} & \color{white}$+$0.8\%
    & {\scriptsize \color{white}65.6\%} & {\scriptsize \color{white}65.0\%} & \color{white}$-$0.6\% \\
  \rowcolor{gray!6}
  \quad p50\,(ms)
    & {\scriptsize 16.51} & {\scriptsize 12.70} & \cellcolor{green!8}1.30$\times$
    & {\scriptsize 11.59} & {\scriptsize 10.15} & \cellcolor{green!8}1.14$\times$
    & {\scriptsize 28.63} & {\scriptsize 23.75} & \cellcolor{green!8}1.21$\times$
    & {\scriptsize 14.55} & {\scriptsize 12.83} & \cellcolor{green!8}1.13$\times$ \\
  \rowcolor{gray!6}
  \quad mean\,(ms)
    & {\scriptsize 18.80} & {\scriptsize 15.57} & \cellcolor{green!8}1.21$\times$
    & {\scriptsize 15.68} & {\scriptsize 14.65} & \cellcolor{green!8}1.07$\times$
    & {\scriptsize 34.24} & {\scriptsize 30.15} & \cellcolor{green!8}1.14$\times$
    & {\scriptsize 17.59} & {\scriptsize 15.60} & \cellcolor{green!8}1.13$\times$ \\

  \bottomrule
  \end{tabular*}

  \label{tab:main_results}
\end{table*}

We evaluate \NAME on state-of-the-art models from four families on eight benchmarks (\cref{sec:eval:setup}).
In all cases, Lynx does not require any task-specific profiling or configuration.
Our key findings are:
\squishlist
\item \NAME provides up to 1.30$\times$ lower latency, with \emph{less than 1 percentage point} impact on accuracy. This accuracy bound is respected across model families, architectures, benchmarks and real-world datasets~\cref{eval:main}.  On average, \NAME improves the accuracy through efficient expert remapping suggesting experts in MoEs carry redun-
dant information due to training-time load balancing

\item \NAME reduces MoE inference latency across the most commonly used inference deployments: co-located and disaggregated serving. \NAME also improves system throughput under a given SLO budget by upto 2.1$\times$~(\cref{eval:SLO}) and provides upto 1.25$\times$  lower latency on real-world traces(\cref{eval:real-world}).

\item \NAME boosts the performance of state-of-the-art vision and thinking models by upto 1.21$\times$ thereby extending to latest generation of models without any modifications~\cref{fig:new-models}.
\item \NAME is complementary to existing approaches. %
\NAME boosts the performance of Fiddler~\cite{Kamahori2024FiddlerCO}, a SOTA offloading technique, by 31\% and LLM compressor's INT4 compression technique, by 10\%~(\cref{eval:quant}).
\squishend

\subsection{Experimental Setup}\label{sec:eval:setup}
\textbf{Models and Experimental Setup.} We evaluate \NAME across models from four families as illustrated in Table~\ref{tab:models-used}. All models use BF16 precision unless otherwise stated.

\noindent \textbf{Code and math benchmarks: }We show performance of \NAME
on four SOTA text generation benchmarks in code and math, namely: GSM8K\cite{Cobbe2021TrainingVT}, HumanEval ~\cite{Chen2021EvaluatingLL}, MBPP~\cite{austin2021programsynthesislargelanguage} and Minerva Math (Algebra)~\cite{lewkowycz2022solvingquantitativereasoningproblems}. These tasks have questions of varying difficulty, input/output lengths. 

\noindent \textbf{Vision and reasoning benchmarks: } 
We also evaluate \NAME on multi-modal and reasoning benchmarks, namely ChartQA~\cite{masry2022chartqabenchmarkquestionanswering}, a benchmark on visual reasoning,  MMMU~\cite{yue2024mmmumassivemultidisciplinemultimodal}, for complex reasoning on inputs like images and text. Finally, we include AIME and GPQA which have problems requiring domain knowledge, multi-step reasoning and mathematical insight. 

\noindent \textbf{Real-world traces: } To show the benefits of \NAME on realistic serving scenarios, we evaluate on  conversation and agentic tool traces from Mooncake~\cite{qin2025mooncake} and ShareGPT~\cite{sharegpt_vicuna_unfiltered}.
These real-world traces include real arrival rates, input/output lengths, and reflect realistic prefix-sharing characteristics.

\textbf{Metrics.} We report the median and mean time-per-output-token latency of \NAME compared to the vLLM baseline with all default inference optimizations enabled.
We also measure overall system throughput (tokens/sec) and user throughput (tokens/sec/user). 

\textbf{Accuracy.} We follow task-specific guidelines (accuracy metrics, fewshot examples, chat template, thinking token budget, etc.) from EleutherAI's benchmarking harness and the technical report of respective models.

\textbf{Experimental Setup.} We evaluate on %
NVIDIA H200 GPUs (141~GB memory) connected using SXM NVLink to conduct all our experiments unless otherwise stated. The machine is equipped with 2x AMD EPYC 9554 64-Core CPU (128 cores total) and 1.5~TB DRAM, running Ubuntu 22.04.4 LTS, with NVIDIA driver 560.35.05 and CUDA 12.6. 

\textbf{Parallelism.} We run Mixtral-8x7B, Qwen2-57B and Llama-4-Scout with tensor parallelism degree of two, Qwen3-30B-A3B on on a single GPU due to smaller parameter count, and Deepseek-Coder, Llama-Maverick, and Qwen3-235B with tensor parallel degree of four.

\textbf{Baseline and serving scenarios:} We run all experiments with online serving setting with vLLM version (v0.10.1) using the v1 scheduler. We evaluate both colocated and disaggregated prefill/decode scenarios.

We measure accuracy using Eleuther AI's LM evaluation harness~\cite{eval-harness} with number of concurrent requests set to saturate the request queue. We evaluate \NAME with (a) maximum batch size set to 16 during decode, and (b) without any limit on maximum batch size for datasets with request arrival timestamps. We additionally evaluate tensor parallel and expert parallel configurations to reflect real-world production deployments~\cite{Agrawal2024VidurAL, Stojkovic2024DynamoLLMDL, nvidia_dynamo_2025}.

For the computation breakdown, we capture the kernel-level latencies within each iteration using the Pytorch profiler.

\subsection{Impact on Accuracy \& Latency}
\label{eval:main}

Table~\ref{tab:main_results} compares \NAME with \textit{vllm} model serving across math and coding workloads with prefill/decode colocation. Overall, \NAME achieves lower latency across the board while adhering to the 1\% point accuracy budget, and in several cases even improves accuracy.

\NAME lowers median TPOT by 1.09--1.30$\times$ across all four models and benchmarks. Qwen2-57B achieves the largest median latency reduction, with 1.25--1.30$\times$ lower latency across all four tasks (accuracy within 0.8\%). DeepSeek-Coder-V2 delivers 1.20--1.22$\times$ lower median latency across benchmarks, and improves HumanEval Pass@1 by 1.9\%. Qwen3-30B, the smallest model by active parameters, achieves 1.14--1.22$\times$ lower latency. Mixtral-8$\times$7B shows more modest reduction (1.09--1.18$\times$) due to fewer total experts which limits remapping search space.

\NAME lowers latency for decode-only iterations where \NAME's remapping is applied~(\S\ref{subsec:Phase}). 
Even when we include the compute-dominated chunked-prefill iterations (that also produce decode tokens), \NAME lowers the mean latency across the board: Qwen2-57B achieves 1.08--1.23$\times$, DeepSeek-Coder-V2 achieves 1.08--1.14$\times$, and Qwen3-30B achieves 1.07--1.15$\times$ lower mean latency.

Across all 16 model-benchmark pairs, accuracy deviations remain within 1\%, and \NAME sometimes improves accuracy. DeepSeek-Coder-V2 gains +1.9\% on HumanEval while Mixtral-8x7B-Instruct gains +1.2\% on the same task and +0.8\% on MATH, suggesting experts in MoEs carry redundant information due to training-time load balancing.

\begin{figure}[!ht]
  \centering
  \includegraphics[width=1\columnwidth]{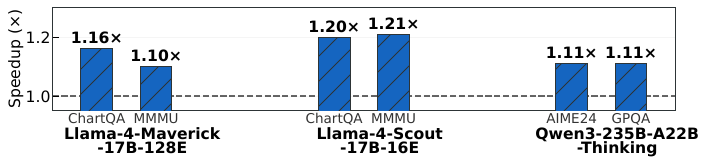}
  \caption{LYNX improves throughput on for SOTA vision and thinking models with no accuracy degradation}
  \label{fig:new-models}
\end{figure}

\paragraph{Vision and Reasoning Tasks.}
As Figure~\ref{fig:new-models} shows, \NAME's latency reduction also provides up to 1.20$\times$ higher throughput on vision models for complex image reasoning tasks (ChartQA and MMMU) on Llama 4.1 Scout and Maverick models, and 1.22$\times$ higher throughput for thinking tasks (AIME24 and GPQA) without any degradation in model accuracy (+0.0 to +3.33\% improvement across tasks) on Qwen 3-235B thinking model.

\paragraph{Disaggregated Serving.} 

\begin{figure}[t]
  \centering
  \includegraphics[width=0.4\columnwidth]{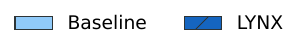}
  \\[2pt]
  \begin{subfigure}[b]{\columnwidth}
    \centering
    \includegraphics[width=\textwidth]{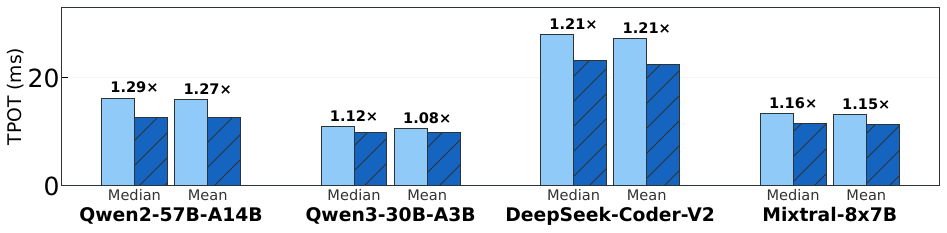}
    \caption{HumanEval}
    \label{fig:disagg-he}
  \end{subfigure}
  \\[2pt]
  \begin{subfigure}[b]{\columnwidth}
    \centering
    \includegraphics[width=\textwidth]{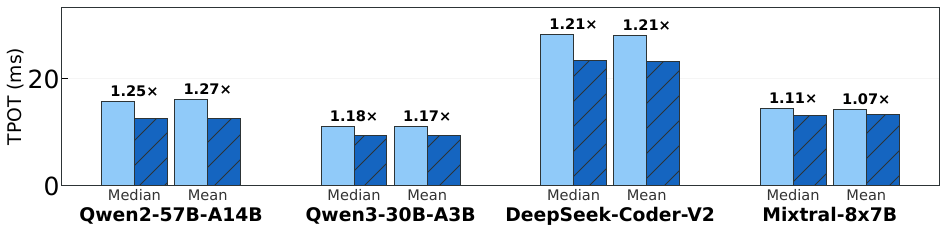}
    \caption{MBPP}
    \label{fig:disagg-mbpp}
  \end{subfigure}
  \\[2pt]
  \begin{subfigure}[b]{\columnwidth}
    \centering
    \includegraphics[width=\textwidth]{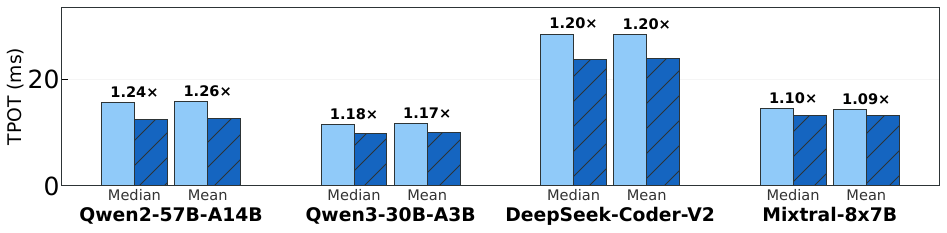}
    \caption{MATH}
    \label{fig:disagg-math}
  \end{subfigure}
  \\[2pt]
  \begin{subfigure}[b]{\columnwidth}
    \centering
    \includegraphics[width=\textwidth]{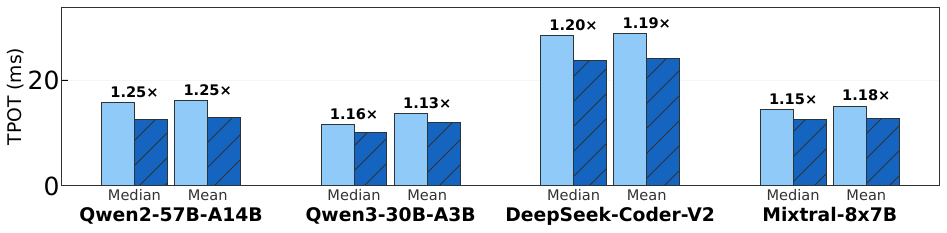}
    \caption{GSM8K}
    \label{fig:disagg-gsm8k}
  \end{subfigure}
  \caption{TPOT latency (ms) when \textsc{Lynx} is applied to a decode-only node
  with disaggregated serving. Speedup shown above each pair.}
  \label{fig:disagg-speedup}
\end{figure}

 Figure~\ref{fig:disagg-speedup} shows \NAME's latency reduction in disaggregated deployment, where prefill and decode run on separate machines, and the decode node processes only decode iterations. As the decode node does not execute chunked-prefill iterations, median and mean latency reductions converge: \NAME lowers both median and mean latency by 1.24--1.35$\times$ on Qwen2-57B, with less than 2\% variation in the mean and median metrics on three of four tasks, and we observe trends for DeepSeek-Coder-V2 (1.19--1.22$\times$ lower TPOT). \NAME lowers Mixtral-8x7B TPOT by 1.15$\times$ on GSM8K, up from 1.13$\times$ in the co-located setting. 
We note that in disaggregated setting, the TPOT reduction for \NAME matches or exceeds median reduction in able~\ref{tab:main_results}.
Finally, absolute latency savings grow with model size: larger models (e.g., Qwen2-57B and DeepSeek-Coder-V2) achieve reductions of up to 6\,ms/token, compared to 1--2\,ms/token for Qwen3-30B.
Thus, \NAME is consistently lower TPOT while being iso-accurate.

\subsection{Improving throughput within budget}
\label{eval:SLO}

\begin{figure}[!h]
  \centering
  \includegraphics[width=0.3\columnwidth]{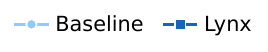}
  \\[2pt]
  \begin{subfigure}[b]{0.48\columnwidth}
    \centering
    \includegraphics[width=\textwidth]{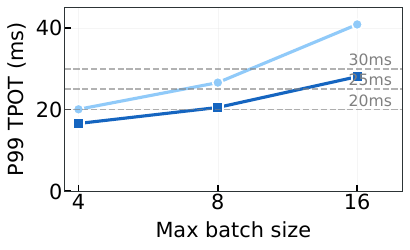}
    \caption{Latency vs.\ batch size}
    \label{fig:qwen2-latency-bs}
  \end{subfigure}
  \hfill
  \begin{subfigure}[b]{0.48\columnwidth}
    \centering
    \includegraphics[width=\textwidth]{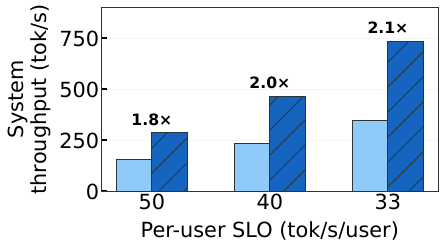}
    \caption{Throughput within SLO}
    \label{fig:qwen2-throughput-slo}
  \end{subfigure}
  
  \caption{\NAME is able to achieve higher system throughput while adhering to the same user token throughput (1/(TPOT)}
  \label{fig:qwen2-slo}
\end{figure}

Serving operators such as OpenAI guarantee per-user token generation rates as part of their API contracts~\cite{openai_api_pricing}. These guarantees translate to P99 TPOT deadlines--25\,ms for 40\,tok/s/user--that limit admissible batch size and, in turn, system throughput. Agentic workloads demand tighter deadlines (20\,ms, 50\,tok/s/user)~\cite{li2025adaserveacceleratingmultislollm}, while chatbot-style serving operates at more relaxed targets (30\,ms, 33\,tok/s/user)~\cite{10.5555/3691938.3691949}. Figure~\ref{fig:qwen2-latency-bs} shows P99 TPOT as a function of batch size for Qwen2-57B-A14B-Instruct on ShareGPT~\cite{sharegpt_vicuna_unfiltered}. \NAME consistently lowers P99 TPOT, allowing the system to admit larger batches under the same deadline. At batch size 16, baseline exceeds 40\,ms while \NAME remains under 30\,ms. Figure~\ref{fig:qwen2-throughput-slo} shows the resulting throughput at all three SLO targets: \NAME achieves 1.8$\times$, 2.0$\times$, and 2.1$\times$ higher throughput, respectively, significantly improving system's efficiency.

\begin{figure*}[!h]
  \centering
  \includegraphics[width=0.5\textwidth]{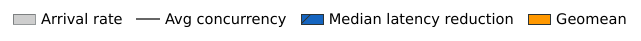}
  \\[2pt]
  \begin{subfigure}[b]{0.4\textwidth}
    \centering
    \includegraphics[width=\textwidth]{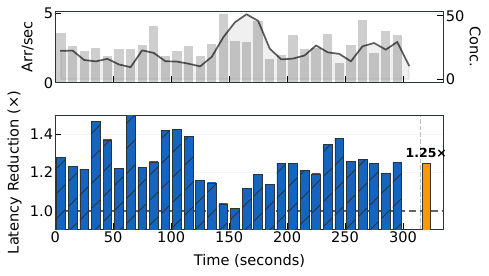}
    \caption{Conversation trace}
    \label{fig:conv-trace}
  \end{subfigure}
  \begin{subfigure}[b]{0.4\textwidth}
    \centering
    \includegraphics[width=\textwidth]{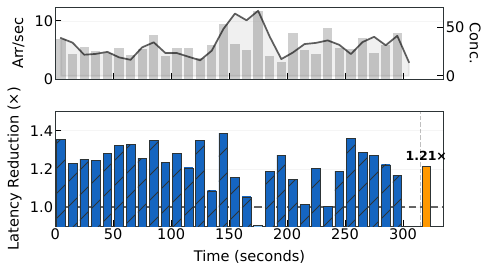}
    \caption{Tool-agent trace}
    \label{fig:toolagent-trace}
  \end{subfigure}
  \\[4pt]
  \caption{LYNX shows significant latency reduction under real-world multi-turn traces for conversation and agentic workloads.}
  \label{fig:aiperf-traces}
\end{figure*}

\begin{figure}[t]
  \centering
  \includegraphics[width=0.5\columnwidth]{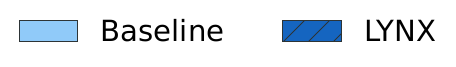}
  \\[2pt]
  \includegraphics[width=0.7\columnwidth]{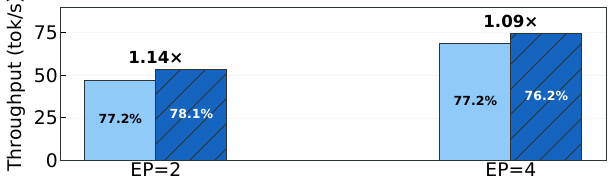}
  \caption{Qwen2-57B-A14B GSM8K: Baseline vs.\ LYNX with expert parallelism.
  Values inside bars show exact-match accuracy (\%).}
  \label{fig:gsm8k-ep}
\end{figure}

\subsection{Expert Parallelism}

\NAME compliments expert parallelism (EP), which distributes experts across GPUs, instead of sharding them (as done in tensor parallelism). Figure~\ref{fig:gsm8k-ep} shows throughput and accuracy on GSM8K for Qwen2-57B-A14B-Instruct at TP=2,EP=2 and TP=4,EP=4. \NAME improves throughput by 1.14$\times$ at EP=2 and 1.09$\times$ at EP=4, while keeping accuracy within 1\% at both configurations. 
The gains reduce at EP=4 because distributing experts across more GPUs reduces the experts per GPU, leaving fewer placement configurations for \NAME to exploit. Nonetheless, 
\NAME provides additive throughput on top of EP scaling, showcasing \NAME's generalizibility to any parallelism technique (TP, EP, PP).
\subsection{LYNX on Real-World Traces}
\label{eval:real-world}
\NAME delivers consistent latency reductions under production-representative workloads with bursty, time-varying arrival rates. We replay two traces: a conversational trace with short, frequent exchanges, and a tool-agent trace with longer requests and higher concurrency~\cite{qin2025mooncake}. Figure~\ref{fig:conv-trace} shows the conversational trace over a 300-second window, where arrival rates fluctuate between 0 and 5 requests/second with concurrency peaking near 50; \NAME lowers median ITL by 1.25$\times$. Figure~\ref{fig:aiperf-traces} shows the tool-agent trace, where arrival rates reach 10 requests/second with similar peak concurrency. \NAME lowers median ITL by 1.21$\times$. In both traces, the TPOT reduction remains temporally stable, including during bursts where concurrency spikes and batch compositions shift rapidly. No offline profiling or trace-specific configuration is required: \NAME's per-batch remapping adapts to shifting arrival patterns, batch sizes, and expert distributions entirely at serving time. These results show that our gains in Table \ref{tab:main_results} carry over to realistic settings.

\subsection{Enhancing Offloading Techniques} %
\label{eval:offloading}
\begin{figure}[!htb]
\centering
\includegraphics[width=0.5\columnwidth]{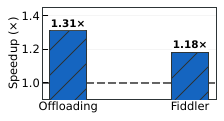}
\caption{\NAME enhances existing offloading techniques}
\label{fig:offload}
\end{figure}
Offloading techniques~\cite{Yu2025fMoEFE, Xue2024MoEInfinityAE,Hwang2023PregatedMA, Kamahori2024FiddlerCO} reduce GPU memory requirements, and \NAME complements such techniques as it enables fetching less expert weights just-in-time from slower memory tiers (like CPU-GPU links) in the critical path.
 
We evaluate two offloading techniques when running Mixtral-8x7B-Instruct on a single A100 GPU with 19\,GB out of 94\,GB offloaded to CPU memory. In conventional offloading, the transfer of model weights is in the critical path, and becomes bottlenecked by PCIe bandwidth, making offloaded inference roughly 50$\times$ slower than on-device inference. \NAME reduces the dominant cost of data transfer of model weights by reducing activated experts per batch. Even at 75\% activated experts, \NAME improves performance over vanilla offloading by $1.31\times$ (\Cref{fig:offload} left).
A recent work, Fiddler~\cite{Kamahori2024FiddlerCO} instead computes offloaded experts on the CPU and transfers the much smaller activation tensors. The critical path thus shifts to expert computation on the CPU, and Fiddler provides a 5$\times$ latency improvement over vanilla offloading. \NAME further enhances Fiddler by reducing CPU-side activated experts by 25\%, yielding an additional $1.18\times$ latency benefit. Overall, \NAME compliments and improves both conventional and state-of-the-art offloading techniques.

\subsection{Enhancing Quantization Techniques} %
\label{eval:quant}

Recent quantization techniques accelerate MoE serving by compressing model weights to FP8 or lower~\cite{he2024demystifyingcompressionmixtureofexpertsunified}, reducing the data transfer cost of expert weights. \NAME is complementary to quantization: it reduces the number of activated experts, further reducing data transfer even for quantized weights.
\begin{figure}[!h]
  \centering
  \includegraphics[width=0.6\columnwidth]{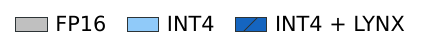}
  \\[1pt]
  \includegraphics[width=0.8\columnwidth]{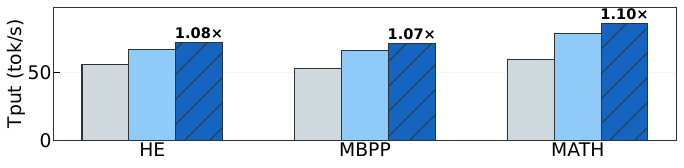}
  \caption{LYNX improves throughput on top of  quantization}
  \label{fig:quant-throughput}
\end{figure}
We evaluate \NAME with Qwen2-57B-A14B-Instruct quantized to INT4 precision using LLM Compressor (GPTQ/AWQ), which is supported by our evaluation framework, vLLM. As shown in Figure~\ref{fig:quant-throughput}, \NAME provides 7–10\% speedup while remaining within the accuracy budget of <1\% (improving MBPP accuracy by 0.8\%) . Other works employ online dynamic pruning on top of extreme quantization~\cite{he2024demystifyingcompressionmixtureofexpertsunified, huang2025mixture}, but their accuracy drops vary drastically on hard tasks such as GSM8K and HumanEval, so we exclude them from our analysis. Thus, we believe \NAME is applicable to any quantization scheme with similar benefits.

\subsection{Impact of Batch Size \& Sequence Length}
\label{eval:scaling}
\begin{figure}[t]
  \centering
  \includegraphics[width=0.4\columnwidth]{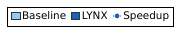}
  \\[2pt]
  \begin{subfigure}[b]{0.48\columnwidth}
    \centering
    \includegraphics[width=\textwidth]{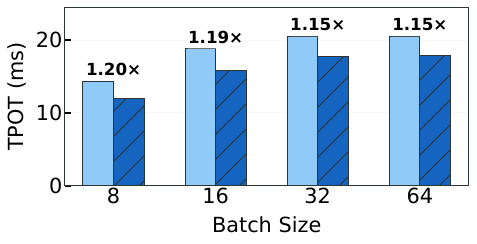}
    \caption{GSM8K}
    \label{fig:batch-scalability-gsm8k}
  \end{subfigure}
  \hfill
  \begin{subfigure}[b]{0.48\columnwidth}
    \centering
    \includegraphics[width=\textwidth]{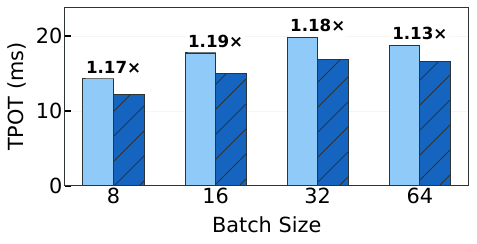}
    \caption{HumanEval}
    \label{fig:batch-scalability-humaneval}
  \end{subfigure}
  \caption{Batch scalability showing robustness of LYNX speedup across different batch sizes and P:D ratios.}
  \label{fig:batch-scalability}
\end{figure}

Figure~\ref{fig:batch-scalability} reports how \NAME’s gains vary across batch sizes on GSM8K (left) and HumanEval (right). The findings are intuitive: TPOT improvements increase with batch size up to a point, after which they begin to saturate.
On GSM8K, \NAME delivers its strongest benefits at moderate batch sizes, achieving over 1.20$\times$ speedup at batch size 8 and 1.19$\times$ at batch size 16, before stabilizing near 1.15$\times$ at 32 and 64. HumanEval shows the same trend, with speedups climbing to 1.19$\times$ at batch size 16 and remaining in the 1.13–1.18$\times$ range thereafter. Importantly, these improvements hold consistently across benchmarks with different prefill to decode ratios. Overall, these results demonstrate that \NAME achieves robust latency reductions across tasks, with particularly strong benefits at moderate batch sizes before saturating, underscoring its effectiveness in practical deployment regimes.

\subsection{Overheads \& scalability}
\label{subsec:eval_overhead}

\begin{figure}[t]
  \centering
  \includegraphics[width=0.9\columnwidth]{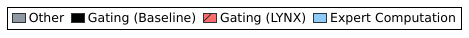}

  \vspace{2pt}

  \begin{subfigure}[b]{0.48\columnwidth}
    \centering
    \includegraphics[width=\linewidth]{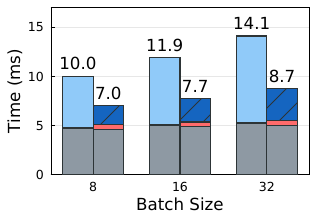}
    \caption{Sequence length 512}
    \label{fig:comp-breakdown-512}
  \end{subfigure}
  \hfill
  \begin{subfigure}[b]{0.48\columnwidth}
    \centering
    \includegraphics[width=\linewidth]{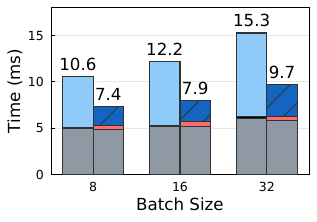}
    \caption{Sequence length 4096}
    \label{fig:comp-breakdown-4096}
  \end{subfigure}

  \caption{Latency breakdown for baseline and LYNX}
  \label{fig:comp-breakdown}
\end{figure}

Figure~\ref{fig:comp-breakdown} profiles the breakdown of end-to-end latency across expert and non-expert components. Due to the highly optimized implementation using 4 fused kernels (\cref{sec:implementation}), the overhead of LYNX’s kernels is less than 4\% overall, making it lightweight and allowing its latency savings from expert reduction to be fully realized in practice. This already minimal overheads can be improved further by tuning kernel dispatch parameters for variable number of activated experts, an optimization that current LLM serving engines ignore.

\section{Related Work}\label{sec:related}

\noindent\textbf{Static Expert Pruning.} Early approaches create smaller MoE variants by static pruning~\cite{Li2023MergeTC} \cite{lasby2026reapexpertspruningprevails}, optionally followed by fine-tuning \cite{Muzio2024SEERMoESE} or distillation ~\cite{xie2024moeprunerpruningmixtureofexpertslarge} to mitigate accuracy loss. For MoEs with a large number of experts, exhaustively navigating the search space is prohibitively expensive, inspiring ~\cite{Liu2024EfficientEP, dong2025domain,Zhang2025MoNERR} to use evolutionary search.

\noindent\textbf{Dynamic Expert Selection.} Many dynamic expert reduction techniques are limited to reducing the number of experts activated per token but not per batch, not realizing latency benefits ~\cite{Lu2024NotAE, he2024demystifyingcompressionmixtureofexpertsunified}. Recent works like \cite{wu2026sere} explore runtime expert selection based on pairwise expert similarity determined from a calibration dataset. \NAME obviates the need for a calibration dataset and, therefore, prior knowledge of the workload's input distribution or any fine-tuning.

\noindent\textbf{Model Compression.} Compression techniques adapt pruning and quantization for MoEs~\cite{he2024demystifyingcompressionmixtureofexpertsunified, huang2025mixture, Chen2025EACMoEEA,Zhang2025MoNERR}. FP16/FP8 quantization~\cite{xiao2023smoothquant,lin2023awq} can reduce memory footprint by 50-75\%, at the expense of accuracy. Large models exceeding GPU HBM in size rely on quantized models with FP8 and even more extreme width quantization like 2-bit and 3-bit width. We show that \NAME complements (4bit and 8bit) quantization and such compression approaches for latency-sensitive deployments. 

\noindent\textbf{Expert Placement and Execution.} Several systems focus on efficient expert distribution across devices. GShard~\cite{Lepikhin2020GShardSG} introduces expert parallelism. Recent works ~\cite{Huang2023TowardsMD, Chen2022TaskSpecificEP, Xue2024MoEInfinityAE,Eliseev2023FastIO, DeepSeekAI2024DeepSeekV3TR} optimize expert placement in constrained and high throughput serving scenarios. \NAME operates at the router and therefore can adapt to load-aware expert placement techniques.

\noindent\textbf{Pipelining and Overlapping} Klotski~\cite{fang2025klotskiefficientmixtureofexpertinference} addresses bubbles in the pipelined inference of MoEs and due to variance in IO and execution latencies per layer by using expert activation patterns to inform multi-batching. Reordering experts in the execution schedule enables overlapping computation with communication, reducing bubbles in the inference pipeline.
\noindent\textbf{Speculative Decoding}
To accelerate low latency LLM serving, multiple tokens are generated using a \textit{draft} model which the large \textit{target} LLM can verify. Since speculative decoding would activate more experts, \emph{verify} stage can be costly. \NAME can work on top of speculative decoding, to lower overhead of verification during speculation.~\cite{liu2024optimizingspeculativedecodingserving}

\section{Conclusion}\label{sec:conclusion}
We present \NAME, the first workload agnostic system to our best knowledge that accelerates MoE inference by addressing the memory bandwidth bottleneck. The key to \NAME's ability is its lightweight run-time batch-level dynamic expert remapping through \emph{AffinityBinning} technique built on the principles revealed in this work, that reduces the number of experts activated per-batch. \NAME achieves upto 1.3$\times$ lower latency on state-of-the-art models from four popular families on six benchmarks while incurring at most 1\% accuracy loss, and even improving accuracy on average. Further, \NAME's complementary nature makes it easy to apply atop existing techniques to further improve their performance.

\phantomsection
\label{EndOfPaper}

{
  \newpage
  \bibliographystyle{ACM-Reference-Format}
  \balance

}

\end{document}